\documentclass{article} 
\usepackage{nips11submit_e,times}
\usepackage{bm,amsmath,amsfonts}
\usepackage{multirow}
\usepackage{epsfig}
\newcommand\x{\mathbf{x}}
\renewcommand\v{\mathbf{v}}
\renewcommand\u{\mathbf{u}}

\newcommand\diag{\mathrm{diag}}

\renewcommand\d{\mathbf{d}}

\newcommand\h{\mathbf{h}}

\renewcommand\Re{\mathbb{R}}
\newcommand\e{\mathbf{e}}

\newcommand\g{\mathbf{g}}
\newcommand\C{\mathbf{C}}

\newcommand\J{\mathbf{J}}

\newcommand\I{\mathbf{I}}
\newcommand\F{\mathbf{F}}
\newcommand\G{\mathbf{G}}

\newcommand\V{\mathbf{V}}
\newcommand\p{\mathbf{p}}
\renewcommand\H{\mathbf{H}}

\renewcommand\a{\mathbf{a}}

\renewcommand\C{\mathbf{C}}
\newcommand\D{\mathbf{D}}

\newcommand\X{\mathbf{X}}
\newcommand\W{\mathbf{W}}
\newcommand\w{\mathbf{w}}

\newcommand\Y{\mathbf{Y}}
\newcommand\y{\mathbf{y}}

\newcommand\btheta{{\bm{\theta}}}

\usepackage{algorithm}
\usepackage{algorithmic}
\renewcommand\algorithmiccomment[1]{ {\footnotesize // #1 } } 

\title{Krylov Subspace Descent for Deep Learning}

\author{
Oriol Vinyals \\
Department of Computer Science\\
U. C. Berkeley \\
Berkeley, CA 94704
\And
Daniel Povey \\
Microsoft Research \\
One Microsoft Way \\
Redmond, WA  98052
}

\nipsfinalcopy 

\begin{document}

\maketitle

\begin{abstract}
In this paper, we propose a second order optimization method to learn models where both the dimensionality of the parameter space and the number of training samples is high.  In our method, we construct on each iteration a Krylov subspace formed by the gradient and an approximation to the Hessian matrix, and then use a subset of the training data samples to optimize over this subspace.  As with the Hessian Free (HF) method of~\cite{martens2010deep}, the Hessian matrix is never explicitly constructed, and is computed using a subset of data.  In practice, as in HF, we typically use a positive definite substitute for the Hessian matrix such as the Gauss-Newton matrix.  We investigate the effectiveness of our proposed method on deep neural networks, and compare its performance to widely used methods such as stochastic gradient descent, conjugate gradient descent and L-BFGS, and also to HF. Our method leads to faster convergence than either L-BFGS or HF, and generally performs better than either of them in cross-validation accuracy.  It is also simpler and more general than HF, as it does not require a positive semi-definite approximation of the Hessian matrix to work well nor the setting of a damping parameter.  The chief drawback versus HF is the need for memory to store a basis for the Krylov subspace.
\end{abstract}

\section{Introduction}
\label{sec:intro}

Many algorithms in machine learning and other scientific computing fields rely on optimizing a function with respect to a parameter space.  In many cases, the objective function being optimized takes the form of a sum over a large number of terms that can be treated as identically distributed: for instance, labeled training samples. Commonly, the problem that we are trying to solve consists of minimizing the negated log-likelihood:

\begin{equation}
    f(\btheta) = -\log(p(\Y|\X;\btheta)) = -\sum_{i=1}^N\log(p(\y_i|\x_n;\btheta)) \label{eqn:objf}
\end{equation}
where $(\X,\Y)$ are our observations and labels respectively, and $p$ is the posterior probability of our labels which is modeled by a deep neural network with parameters $\btheta$. In this case it is possible to use subsets of the training data to obtain noisy estimates of quantities such as gradients; the canonical example of this is Stochastic Gradient Descent (SGD).

The simplest reference point to start from when explaining our method is Newton's method with line search, where on iteration $m$ we do an update of the form:

\begin{equation}
   \theta_{m+1} = \theta_m - \alpha \H_m^{-1} \g_m,  \label{eqn:newton}
\end{equation}
where $\H_m$ and $\g_m$ are, respectively, the Hessian and the gradient on iteration $m$ of the objective function~\eqref{eqn:objf}; here, $\alpha$ would be chosen to minimize~\eqref{eqn:objf} at $\theta_{m+1}$.  For high dimensional problems it is not practical to invert the Hessian; however, we can efficiently approximate~\eqref{eqn:newton} using only multiplication by $\H_m$, by using the Conjugate Gradients (CG) method with a truncated number of iterations.  In addition, it is possible to multiply by $\H_m$ without explicitly forming it, using what is known as the ``Pearlmutter trick''~\cite{pearlmutter1994fast} (although it was known to the optimization community prior to that; see~\cite[Chapter 8]{Nocedal2006NO}) for multiplying an arbitrary vector by the Hessian; this is described for neural networks but is applicable to quite general types of functions.  This type of optimization method is known as ``truncated Newton'' or ``Hessian-free inexact Newton''~\cite{Morales00enrichedmethods}.  In~\cite{byrd2011use}, this method is applied but using only a subset of data to approximate the Hessian $\H_m$.  A more sophisticated version of the same idea was described in the earlier paper~\cite{martens2010deep}, in which preconditioning is applied, the Hessian is damped with the unit matrix in a Levenberg-Marquardt fashion, and the method is extended to non-convex problems by substituting the Gauss-Newton matrix for the Hessian. We will discuss the Gauss-Newton matrix and its relationship with the Hessian in Section~\ref{sec:gn}.

Our method is quite similar to the one described in~\cite{martens2010deep}, which we will refer to as Hessian Free (HF).  We also multiply by the Hessian (or Gauss-Newton matrix) using the Pearlmutter trick on a subset of data, but on each iteration, instead of approximately computing $(\H_m + \lambda \I)^{-1} \g_m$ using truncated CG, we compute a basis for the Krylov subspace spanned by $\g_m, \H_m \g_m, \ldots \H_m^{K-1} \g_m$ for some $K$ fixed in advance (e.g. $K=20$), and numerically optimize the parameter change within this subspace, using BFGS to minimize the original nonlinear objective function measured on a subset of the training data.  It is easy to show that, for any $\lambda$, the approximate solution to $\H_m + \lambda \I$ found by $K$ iterations of CG will lie in this subspace, so we are in effect automatically choosing the optimal $\lambda$ in the Levenburg-Marquardt smoothing method of HF (although our algorithm is free to choose a solution more general than this).  We note that both our method and HF use preconditioning, which we have glossed over in the discussion above.  Compared with HF, the advantages of our method are:
\begin{itemize}
  \item Greater simplicity and robustness: there is no need for heuristics to initialize and update the smoothing value $\lambda$.
  \item Generality: unlike HF, our method can be applied even if $\H$ (or whatever approximation or substitute we use) is not positive semidefinite.
  \item Empirical advantages: our method generally seems to work better than HF in both optimization speed and classification performance.
\end{itemize}
The chief disadvantages versus HF are:
\begin{itemize}
  \item Memory requirement: we require storage of $K$ times the parameter dimension to store the subspace.
  \item Convergence properties: the use of a subset of data to optimize over the subspace will prevent convergence to an optimum.
\end{itemize}
Regarding the convergence properties: we view this as more of a theoretical than a practical problem, since for typical setups in training deep networks the residual parameter noise due to the use of data subsets would be far less than that due to overtraining.


Our motivation for the work presented here is twofold: firstly, we are interested in large-scale non-convex optimization problems where the parameter dimension and the number of training samples is large and the Hessian has large condition number.   We had previously investigated quite different approaches based on preconditioned SGD to solve an instance of this type of optimization problem (our method could be viewed as an extension to~\cite{le2007topmoumoute}), but after reading~\cite{martens2010deep} our interest switched to methods of the HF type.  Secondly, we have an interest in deep neural nets, particularly to solve problems in speech recognition, and we were intrigued by the suggestion in~\cite{martens2010deep} that the use of optimization methods of this type might remove the necessity for pretraining, which would result in a welcome simplification.  Other recent work on the usefulness of second order methods for deep neural networks includes~\cite{GlorotAISTATS2010,NgICML11}.

\section{The Hessian matrix and the Gauss-Newton matrix}
\label{sec:gn}

The Hessian matrix $\H$ (that is, the matrix of second derivatives w.r.t. the parameters) can be used in HF optimization whenever it is guaranteed positive semidefinite, i.e. when minimizing functions that are convex in the parameters.  For non-convex problems, it is possible to substitute a positive definite approximation to the Hessian.  One option is the Fisher information matrix,
\begin{equation}
  \F = \sum_i \g_i \g_i^T,
\end{equation}
where indices $i$ correspond to samples and the $\g_i$ quantities are the gradients
for each sample.  This is a suitable stand-in for the Hessian because
it is in a certain sense dimensionally the same, i.e. it changes the same way under
transformations of the parameter space.  If the model can be interpreted as producing
a probability or likelihood, it is possible under certain assumptions (including model
correctness) to show that close to convergence, the Fisher and Hessian matrices have
the same expected value.  The use of the Fisher matrix in this way is known as
Natural Gradient Descent~\cite{Amari:1998:NGW:287476.287477}; in~\cite{le2007topmoumoute},
a low-rank approximation of the Fisher matrix was used instead.
Another alternative that has less theoretical justification
but which seems to work better in practice in the case of neural networks is the
Gauss-Newton matrix, or rather a
slight generalization of the Gauss-Newton matrix that we will now describe.

\subsection{The Gauss-Newton matrix}

The Gauss-Newton matrix is defined when we have a function (typically nonlinear) from a
vector to a vector, $f: \Re^n \rightarrow \Re^m$.  Let the Jacobian of this function be
$\J \in \Re^{m \times n}$, then the Gauss-Newton matrix is $\G = \J^T \J$,
with $\G \in \Re^{n\times n}$.  If the problem is least-squares on the output
of $f$, then $\G$ can be thought of as one term
in the Hessian on the input to $f$.  In its application to neural-network training,
for each training example we consider the network as a nonlinear function from the
neural-network parameters $\btheta$ to the output of the network, with the neural-network input
treated as a constant.  As in~\cite{schraudolph}, we
generalize this from least squares to general convex error functions by using the
expression $\J^T \H \J$, where $\H$ is the (positive semidefinite) second derivative
of the error function w.r.t. the neural network output.  This
can be thought of as the part of the Hessian that remains after ignoring the
nonlinearity of the neural-network in the parameters.  In the rest of this document,
following~\cite{martens2010deep} we will refer to this matrix $\J^T \H \J$ simply as
the Gauss-Newton matrix, or $\G$, and depending on the context, we may actually be
referring to the summation of this expression over a number of neural-network training
samples.

\subsection{Efficiently multiplying by the Gauss-Newton matrix}

As described in~\cite{schraudolph}, it is possible to efficiently multiply a
vector by $\G$  using a version of the ``Pearlmutter trick''; the algorithm is similar
in spirit to backprop and we give it as Algorithm~\ref{alg:gn}.  Our notation and our derivation for this
algorithm differ from~\cite{pearlmutter1994fast,schraudolph}, and we will explain this briefly;
we find our approach easier to follow.  The idea is this: we first imagine
that we are given a parameter set $\btheta$, and two vectors $\btheta_1$
and $\btheta_2$ which we interpret as directions in parameter space; we then
write down an algorithm to compute the scalar $s = \btheta_2^T \G \btheta_1$.
Assume the neural-network input is given and fixed;
let $\v$ be the network output, and write it as $\v(\btheta)$ to emphasize the
dependence on the parameters, and then let $\v_1$ be defined as
\begin{equation}
  \v_1 =  \lim_{\alpha \rightarrow 0} \frac{1}{\alpha} \v(\btheta + \alpha \btheta_1) - \v(\btheta), \label{eqn:v1}
\end{equation}
so that $\v_1 = \J \btheta_1$.  We define $\v_2$ similarly.  These can both
be computed in a modified forward pass through the network.  Then, if $\H$ is the
Hessian of the error function in the output of the network (taken at parameter value $\btheta$),
$s$ is given by
\begin{equation}
  s = \v_2^T \H \v_1, \label{eqn:s:1}
\end{equation}
since $\v_2^T \H \v_1 = \btheta_2^T \J^T \H \J \btheta_1 = \btheta_2^T \G \btheta_1$.
The Hessian $\H$ of the error
function would typically not be constructed as a matrix, but we would compute~\eqref{eqn:s:1}
given some analytic expression for $\H$.  Suppose we have written down the algorithm
for computing $s$ (we have not done so here because of space constraints).  Then
we treat $\btheta_1$ as a fixed quantity, but compute the derivative of $s$ w.r.t.
$\btheta_2$ (taking $\btheta_2$ around zero for convenience).  This derivative equals
the desired product $\G \btheta_1$.  Taking the derivative of a scalar
w.r.t. the input to an algorithm can be done in a mechanical fashion via ``reverse-mode''
automatic differentiation through the algorithm, of which neural-net backprop is a special
case.  This is how we obtained Algorithm~\ref{alg:gn}.
In the algorithm we denote the  derivative of $s$ w.r.t. a quantity $x$ by $\hat{x}$, i.e.
by adding a hat.  We note that in this algorithm, we have a ``backward pass'' for
quantities with subscript 2,
which did not appear in the forward pass, because
they were zero (since we take $\btheta_2 = 0$) and we optimized them out.

Something to note here is that when the linearity of the last layer is softmax and the error
is negated cross-entropy (equivalently negated log-likelihood, if the label is known),
we actually view the softmax nonlinearity as part of the error function.  This is a
closer approximation to the Hessian, and it remains positive semidefinite.

To explain the notation of Algorithm~\ref{alg:gn}: $\h^{(i)}$ is the input
to the nonlinearity of the $i$'th layer and $\v^{(i)}$ is the output;
$\odot$ means elementwise multiplication; $\phi^{(i)}$ is the nonlinear function of
the $i$'th layer, and when we apply it
to vectors it acts elementwise; $\W^{(1)}$ is the neural-network weights for the
first layer (so $\h^{(1)} = \W^{(1)} \v^{(0)}$, and so on); we use the subscript
$1$ for quantities that represent how quantities change when we move the parameters
in direction $\btheta_1$ (as in~\eqref{eqn:v1}).
The error function is written as ${\cal E}(\v^{(L)}, y)$ (where $L$ is the last layer),
and $y$, which may be a discrete value, a scalar or a vector, represents the
supervision information the network is trained with.  Typically ${\cal E}$ would
represent a squared loss or negated cross-entropy.
In the squared-loss case, the quantity $\frac{\partial^2}{\partial \v^2} {\cal E}(\v^{(L)}, y)$
in Line~\ref{line:e} of Algorithm~\ref{alg:gn} is just the unit matrix.  The other
case we deal with here is negated cross entropy.  As mentioned above, we include
the soft-max nonlinearity in the error function,
treating the elements of the output layer $\v^{(L)}$ as unnormalized log probabilities.  If
the elements of $\v^{(L)}$ are written as $v_j$ and  we let $\p$ be the vector of
probabilities, with $p_j = \exp(v_j) / \sum_i \exp(v_i)$, then the
matrix of second derivatives is given by
\begin{equation}
\frac{\partial^2}{\partial \v^2} {\cal E}(\v^{(L)}, y) = \diag(\p) - \p \p^T .
\end{equation}

\begin{algorithm}
    \caption{Compute product $\hat{\btheta}_2 = \G \btheta_1$: MultiplyG$(\btheta, \btheta_1, \x, y)$ }
    \label{alg:gn}
    \begin{algorithmic}[1] 
        \STATE \algorithmiccomment{ Note, $\btheta = ( \W^{(1)}, \W^{(2)}, \ldots )$ and $\btheta_1 = ( \W_1^{(1)}, \W_2^{(2)}, \ldots )$. }
        \STATE $\v^{(0)} \gets \x$
        \STATE $\v_1^{(0)} \gets {\mathbf 0}$
        \FOR { $l = 1 \ldots L$ }
           \STATE $\h^{(l)} \gets \W^{(l)} \v^{(l-1)}$
           \STATE $\h_1^{(l)} \gets \W^{(l)} \v_1^{(l-1)} + \W_1^{(l)} \v^{(l-1)}$
           \STATE $\v^{(l)} \gets \phi^{(l)}(\h^{(l)})$
           \STATE $\v_1^{(l)} \gets {\phi'}^{(l)}(\h^{(l)}) \odot \h_1^{(l)}$
        \ENDFOR
        \STATE $\hat{\v}_2^{(L)} \gets \frac{\partial^2}{\partial \v^2} {\cal E}(\v^{(L)}, y) \v_1^{(L)}$ \label{line:e}
        \FOR { $l = L \ldots 1$ }
           \STATE $\hat{\h}_2^{(l)} \gets \hat{\v}_2^{(l)} \odot {\phi'}^{(l)}(\h^{(l)})$
           \STATE $\hat{\v}_2^{(l-1)} \gets \left.\W^{(l)}\right.^T \hat{\h}_2^{(l)}$
           \STATE $\hat{\W}_2^{(l)} \gets \hat{\h}_2^{(l)} \left.\v^{(l-1)}\right.^T$
        \ENDFOR
        \RETURN $\hat{\btheta}_2 \equiv \left( \hat{\W}_2^{(1)}, \ldots, \hat{\W}_2^{(L)} \right)$
     \end{algorithmic}
\end{algorithm}

\section{Krylov Subspace Descent: overview}
\label{sec:overview}
Now we describe our method, and how it relates to Hessian Free (HF) optimization.
The discussion in the previous section (on the Hessian versus Gauss-Newton matrix) is orthogonal
to the distinction between KSD and HF, because either method can use any Hessian substitute, with the
proviso that our method can use the Hessian even when it is not positive definite.

In the rest of this section we will use $\H$ to refer to either the Hessian or a substitute such
as $\G$ or $\F$.  In \cite{martens2010deep} and the work we describe here, these matrices are approximated using a subset of data samples.
In both HF and KSD, the whole computation is preconditioned using the diagonal of $\F$ (since this is
easy to compute); however, in the discussion below we will gloss over this preconditioning.
In HF, on each iteration the CG algorithm is used to approximately compute
\begin{equation}
   \d = - (\H + \lambda \I)^{-1} \g,
\end{equation}
where $\d$ is the step direction, and $\g$ is the gradient.  The step size is determined by a
backtracking line search.  The value of $\lambda$ is kept updated by Levenburg-Marquardt style
heuristics.  Other heuristics are used to control the stopping of the CG iterations.  In addition,
the CG iterations for optimizing $\d$ are not initialized from zero (which would be the natural
choice) but from the previous value of $\d$; this loses some convergence guarantees but seems to
improve performance, perhaps by adding a kind of momentum to the updates.

In our method (again glossing over preconditioning), we compute a basis for the subspace
spanned by $\{ \g, \H \g, \ldots, \H^{K-1} \g, \d_{\mathrm{prev}} \}$, which is the Krylov
subspace of dimension $K$, augmented with the previous search direction.  We then optimize the objective function over this subspace using BFGS, approximating
the objective function using a subset of samples.



\section{Krylov Subspace Descent in detail}

In this section we describe the details of the KSD algorithm, including the
preconditioning.

For notation purposes: on iteration $n$ of the overall optimization we will write
the training data set used to obtain the gradient as ${\cal A}_n$ (which is
always the entire dataset in our experiments); the set used to compute the Hessian
or Hessian substitute as ${\cal B}_n$; and the set used for BFGS
optimization over the subspace, as ${\cal C}_n$.
For clarity when dealing with multiple subset sizes, we will typically normalize all
quantities by the number of samples: that is, objective function values, gradients,
Hessians and the like will always be divided by the number of samples in the
set over which they were computed.

On each iteration we will compute a diagonal preconditioning matrix $\D$ (we omit
the subscript $n$).  $\D$ is expected to be a rough approximation to the Hessian.
In our experiments, following~\cite{martens2010deep}, we set $\D$ to the diagonal
of the Fisher matrix computed over ${\cal A}_n$.
To precondition, we define a new variable $\tilde{\btheta} = \D^{1/2} \btheta$, compute the Krylov
subspace in terms of this variable, and convert back to the ``canonical'' co-ordinates.
The result is the subspace spanned by the vectors
\begin{equation}
 \left\{ (\D^{-1} \H)^k \D^{-1} \g, 0\leq k < K  \right\} \label{eqn:subs}
\end{equation}
We adjoin the previous step direction $\d_\mathrm{prev}$ to this, and it becomes the
subspace we optimize over with BFGS.  The algorithm to compute an orthogonal
basis for the subspace, and the Hessian (or Hessian substitute) within it, is given as
Algorithm~\ref{alg:proj}.

\begin{algorithm}[h]
\caption{Construct basis $\V = \left[\v_1,\ldots,\v_{K+1}\right]$ for the subspace, and the Hessian (or substitute) $\bar{\H}$ in the co-ordinates of the subspace.}
\label{alg:proj}
 \begin{algorithmic}[1] 
    \STATE $\v_1 \gets \D^{-1} \g$
    \STATE $\v_1 \gets \frac{1}{\sqrt{\v_1^T \v_1}} \v_1$
    \FOR { $k = 1 \ldots K+1$ }
      \STATE $\w \gets \H \v_k$  \algorithmiccomment{If Gauss-Newton matrix, computed with Algorithm~\ref{alg:gn}.}
      \IF { $k < K$ }
        \STATE $\u \gets \D^{-1} \w$ \algorithmiccomment{$\u$ will be $\v_{m+1}$}
      \ELSIF { $k = K$ }
        \STATE $\u \gets \d_\mathrm{prev}$ \algorithmiccomment{Previous search direction; use arbitrary nonzero vector if 1st iter}
      \ENDIF
      \FOR { $j = 1 \ldots k$ }
         \STATE $\bar{h}_{k,j} \gets \w^T \v_j$ \algorithmiccomment{Compute element of reduced-dimension Hessian}
         \STATE $\u \gets \u  -  (\u^T \v_j) \v_j$ \algorithmiccomment{Orthogonalize $\u$}
      \ENDFOR
      \IF { $k \leq K$}
          \STATE $\v_{k+1} \gets \frac{1}{\sqrt{\u^T \u}} \u$ \algorithmiccomment{Normalize length and set next direction.}
      \ENDIF
    \ENDFOR
    \STATE \algorithmiccomment{Now set upper triangle of $\bar{\H}$ to lower triangle.}
 \end{algorithmic}
\end{algorithm}


On each iteration of optimization, after computing the basis $\V$ with Algorithm~\ref{alg:proj}
we do a further preconditioning step within the subspace, which gives us a new, non-orthogonal
basis $\hat{\V}$ for the subspace.  This step is done to help the BFGS converge faster.

\begin{algorithm}[h]
    \caption{Krylov Subspace Descent}
    \label{alg:ksd}
 \begin{algorithmic}[1] 
    \STATE $\d_\mathrm{prev} \gets \e_1$ \algorithmiccomment{or any arbitrary nonzero vector}
    \FOR { $n = 1, 2 \ldots$ }
       \STATE \algorithmiccomment{Sample three sets from training data, ${\cal A}_n$, ${\cal B}_n$ and ${\cal C}_n$.}
       \STATE $\g \gets \frac{1}{ |{\cal A}_n| } \sum_{i \in {\cal A}_n}  \g_i(\btheta)$ \algorithmiccomment{ Get average function gradient over this batch. }
       \STATE Set $\D$ to diagonal of Fisher matrix on ${\cal A}_n$, floored to $\epsilon$ times its maximum.
       \STATE Run Algorithm~\ref{alg:proj} to find $\V$ and $\bar{\H}$ on subset ${\cal B}_n$
       \STATE Let $\hat{\H}$ be the result of flooring the eigenvalues of $\bar{\H}$ to $\epsilon$ times the maximum.
       \STATE Do the Cholesky decomposition $\hat{\H} = \C \C^T$
       \STATE Let $\bar{\V} = \V \C^{-T}$ (do this in-place; $\C^{-T}$ is upper triangular)
       \STATE $\a \gets 0 \in \Re^{K+1}$
       \STATE Find the optimum $\a^*$ with BFGS for about $K$ iterations using the subset ${\cal C}_n$, with objective function measured at $\btheta + \bar{\V}\a$ and gradient $\bar{\V}^T \g$ (where $\g$ is the gradient w.r.t. $\btheta$).
       \STATE $\d_\mathrm{prev} \gets \bar{\V}\a^*$
       \STATE $\btheta \gets \btheta + \d_\mathrm{prev}$
     \ENDFOR
 \end{algorithmic}
\end{algorithm}

The complete algorithm is given as Algorithm~\ref{alg:ksd}.  The most important
parameter is $K$, the dimension of the Krylov subspace (e.g. 20).  The flooring
constant $\epsilon$ is an unimportant parameter; we used $10^{-4}$.  The subset sizes may be important; we recommend that ${\cal A}_n $ should be all of the training data, and ${\cal B}_n$ and ${\cal C}_n$ should each be about $1/K$ of the training data, and disjoint from each other but not from ${\cal A}_n$. This is the subset size
that keeps the computation approximately balanced between the gradient computation, subspace
construction and subspace optimization. Implementations of the BFGS algorithm would typically also have parameters: for instance,
parameters of the line-search algorithm and stopping critiera; however, we expect that
in practice these would not have too much effect on performance because
the algorithm is likely to converge almost exactly (since the subspace dimension and the
number of iterations are about the same).

\section{Experiments}
\label{sec:exp}

To evaluate KSD, we performed several experiments to compare it with SGD and with
other second order optimization methods, namely L-BFGS and HF. We report both
training and cross validation errors, and running time (we terminated the algorithms
with an early stopping rule using held-out validation data).  Our
implementations of both KSD and HF are based on Matlab using
Jacket\footnote{www.accelereyes.com} to perform the expensive matrix operations
on a Geforce GTX580 GPU with 1.5GB of memory.

\subsection{Datasets and models}

Here we describe the datasets that we used to compare KSD to other methods.

\begin{table}
\centering
\begin{tabular}{l|c|c|c|c|c|c}
\hline
Dataset&Train smp.&Test smp.&Input&Output&Model&Task\\
\hline
CURVES&20K&10K&784 (bin.)&784 (bin.)&400-200-100-50-25-5&AE\\
MNIST$_{AE}$&60K&10K&784 (bin.)&784 (bin.)&1000-500-250-30&AE\\
MNIST$_{CL}$&60K&10K&784 (bin.)&10 (class)&500-500-2000&Class\\
MNIST$_{CL,PT}\footnotemark[1]$&60K&10K&784 (bin.)&10 (class)&500-500-2000&Class\\
Aurora&1.2M&100K\footnotemark[2]&352 (real)&56 (class)&512-1024-1536&Class\\
Starcraft&900&100&5077 (mix)&8 (class)&10&Class\\
\hline
\end{tabular}
\caption{Datasets and models used in our setup.}
\label{tab:models}
\end{table}

\begin{itemize}
\item CURVES: Artificial dataset consisting of curves at $28\times28$ resolution. The dataset consists of 20K training samples, and 10K testing samples. We considered an autoencoder network, as in \cite{HinSal06}.
\item MNIST: Single digit vision classification task. The digits are $28\times28$ pixels, with a 60K training, and 10K testing samples. We considered both an autoencoder network, and classification \cite{HinSal06}.
\item Aurora: Spoken digits dataset, with different levels of real noise (airport, train station, ...). We used PLP features and performed classification of 56 English phones. These frame level phone error rates are the ones reported in Table~\ref{tab:results}. Also reported in the text are Word Error Rates, which were produced by using the phone posteriors in a Tandem system, concatenated with standard MFCC to train a Hidden Markov Model with Gaussian Mixture Model emissions. Further details on the setup can be found in \cite{VinyalsICASSP11}.
\item Starcraft: The dataset consists of a real time strategy video game sequences from 1000 games. The goal is to predict the strategy the opponent chose based on a fully observed game sequence after five minutes, and features contain orderings between buildings, presence/absence features, or times that certain buildings were built.
\end{itemize}

The models (i.e. network architectures) for each dataset are summarized in Table~\ref{tab:models}. We tried to explore a wide variety of models covering different sizes, input and output characteristics, and tasks. Note that the error reported for the autoencoder (AE) task is the L2 norm squared between input and output, and for the classification (Class) task is the classification error (i.e. 100-accuracy). The non linearities considered were logistic functions for all the hidden layers except for the ``coding'' layer (i.e. middle layer) in the autencoders, which was linear, and the visible layer for classification, which was softmax.

\footnotetext[1]{For MNIST$_{CL,PT}$ we initialize the weights using pretraining RBMs as in \cite{HinSal06}. In the other experiments, we did not find a significant difference between pretraining and random initialization as in \cite{martens2010deep}.}
\footnotetext[2]{We report both classification error rate on a 100K CV set, and word error rate on a 5M testing set with different levels of noise}

\subsection{Results and discussion}

Table~\ref{tab:results} summarizes our results. We observe that KSD converges faster than HF, and tends to lead to lower generalization error. Our implementation for the two methods is almost identical; the steps that dominate the computation (computing objective functions, gradients and Hessian or Gauss-Newton products) are shared between both and are computed on a GPU.

For all the experiments we used the Gauss-Newton matrix (unless otherwise specified). The dimensionality of the Krylov subspace was set to 20, the number of BFGS iterations was set to 30 (although in many cases the optimization on the projected gradients converged before reaching 30), and an L2 regularization term was added to the objective function. However, motivated by the observation that on CURVES, HF tends to use a large number of iterations, we experimented with a larger subspace dimension of $K=80$ and these are the numbers we report in Table~\ref{tab:results}.

\begin{figure}[ht]
\begin{minipage}[b]{0.5\linewidth}
\centering
\includegraphics[width=\linewidth]{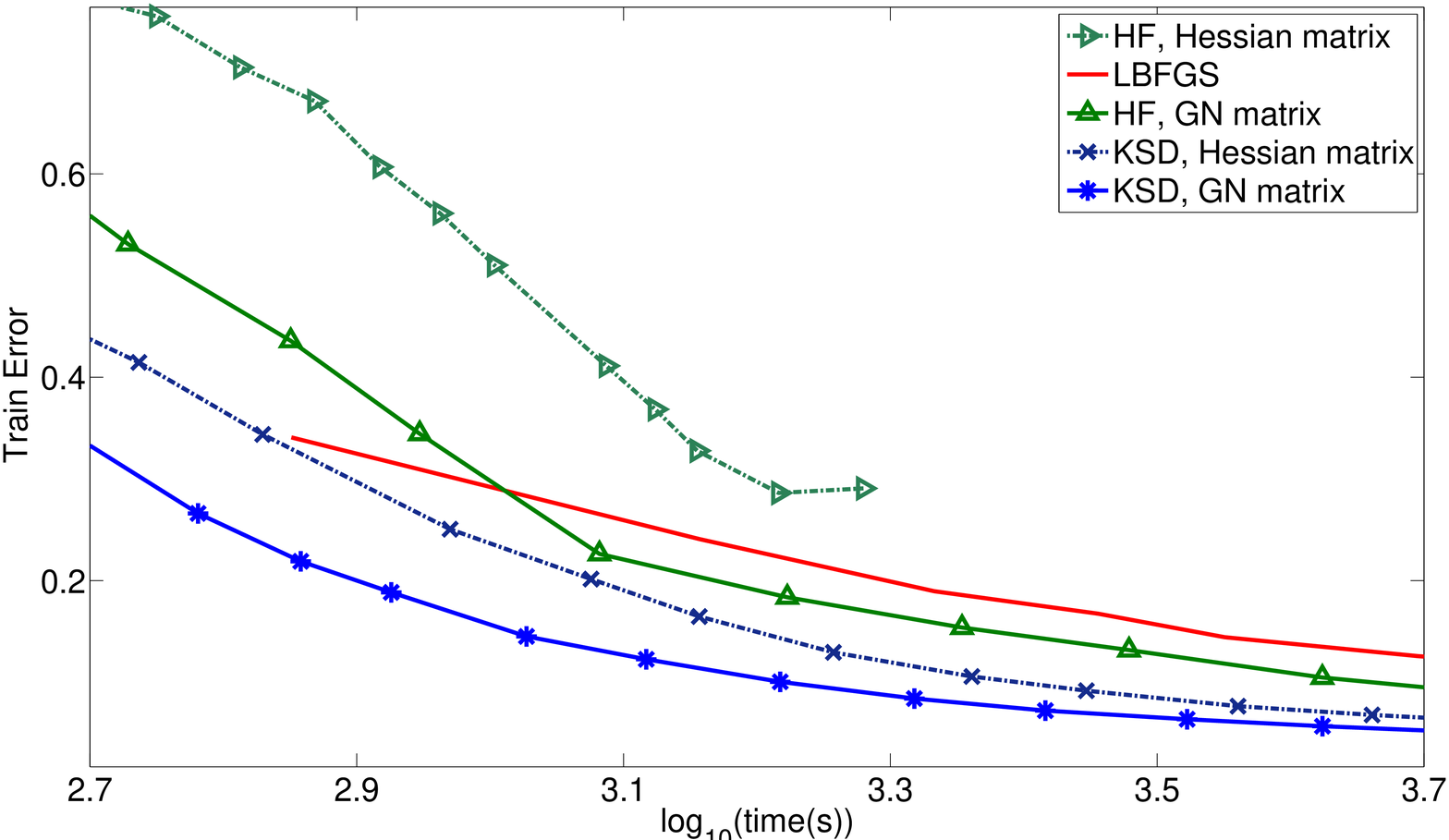}
\caption{Aurora convergence curves for various algorithms.}
\label{fig:aurora}
\end{minipage}
\hspace{0.5cm}
\begin{minipage}[b]{0.5\linewidth}
\centering
\includegraphics[width=\linewidth]{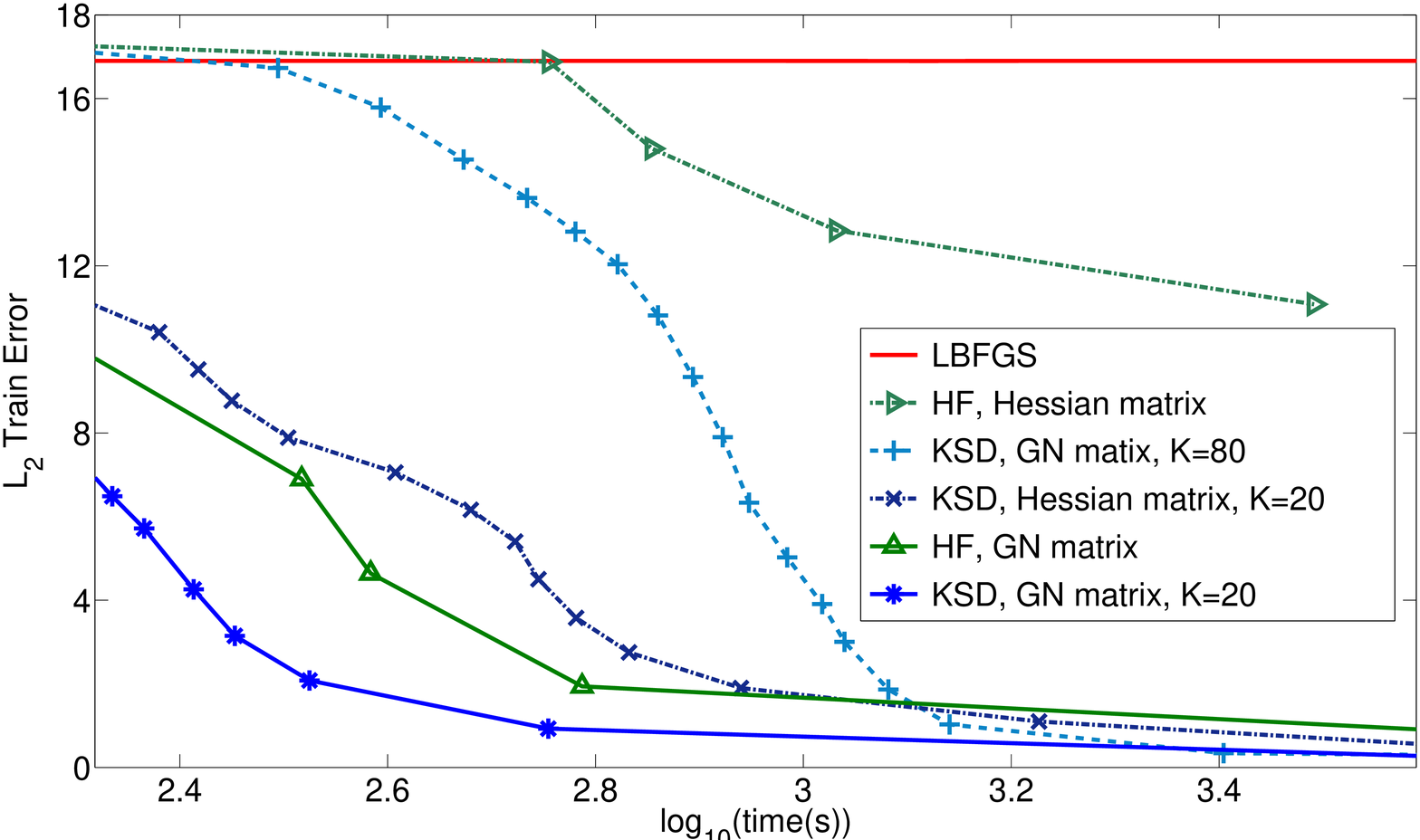}
\caption{CURVES convergence curves for various algorithms.}
\label{fig:curves}
\end{minipage}
\end{figure}

For compatibility in memory usage with KSD, we used a moving window of size 10 for the L-BFGS methods.
We do not report SGD performance in Figures~\ref{fig:aurora} and~\ref{fig:curves} as it was worse
than L-BFGS.

When using HF or KSD, pre-training helped significantly in the MNIST classification task, but not for the other tasks (we do not show the results with pre-training in the other cases; there was no significant difference).  However, when using SGD or CG for optimization (results not shown), pre-training helped on all tasks except Starcraft (which is not a deep network).  This is consistent with the notion put forward in~\cite{martens2010deep} that it might be possible to do away with the need for pre-training if we use powerful second-order optimization methods.  The one exception to this, MNIST, has zero training error when using HF and KSD, which is consistent with a regularization interpretation of pre-training.  This is opposite to the conclusions reached in~\cite{ErhanAISTATS2009} (their conclusion was that pre-training helps by finding a better ``basin of attraction''), but that paper was not using these types of optimization methods.  Our experiments support the notion that when using advanced second-order optimization methods and when overfitting is not a major issue, pre-training is not necessary.  We are not giving this issue the attention it deserves, since the primary focus of this paper is on our optimization method; we may try to support these conclusions more convincingly in future work.

\begin{table}
\centering
\begin{tabular}{l|c|c|c|c|c|c}
\hline
&\multicolumn{3}{|c|}{HF}&\multicolumn{3}{|c}{KSD}\\
\hline
Dataset&Tr. err.&CV err.&Time&Tr. err.&CV err.&Time\\
\hline
CURVES&0.13& \textbf{0.19}&1 &0.17 &0.25 &0.2 \\ 
MNIST$_{AE}$&1.7& 2.7&1 &1.8 &\textbf{2.5} &0.2 \\ 
MNIST$_{CL}$&0\%& 2.01\%&1 &0\% &\textbf{1.70\%} &0.6 \\ 
MNIST$_{CL,PT}$&0\%& 1.40\%&1 &0\% &\textbf{1.29\%} &0.6 \\ 
Aurora&5.1\%& 8.7\%&1 &4.5\% &\textbf{8.1\%} &0.3 \\ 
Starcraft&0\%& 11\%&1 &0\% &\textbf{5\%} &0.7 \\ 
\hline
\end{tabular}
\caption{Results comparing two second order methods: Hessian Free and Krylov Subspace Descent. Time reported is relative to the running time of HF (lower than 1 means faster).}
\label{tab:results}
\end{table}

In Figures~\ref{fig:aurora}~and~\ref{fig:curves}, we show the convergence of KSD and HF with both the Hessian and Gauss-Newton matrices.  HF eventually ``gets stuck'' when using the Hessian; the algorithm was not designed to be used for non-positive definite matrices.  Even before getting stuck, it is clear that it does not work well with the actual Hessian.  Our method also works better with the Gauss-Newton matrix than with the Hessian, although the difference is smaller.  Our method is always faster than HF and L-BFGS.

\section{Conclusion and future work}

In this paper, we proposed a new second order optimization method.  Our approach relies on efficiently computing the matrix-vector product between the Hessian (or a PSD approximation to it), and a vector. Unlike Hessian Free (HF) optimization, we do not require the approximation of the Hessian to be PSD, and our method requires fewer heuristics; however, it requires more memory.

Our planned future work in this direction includes investigating the circumstances under which pre-training is necessary: that is, we would like to confirm our statement that pre-training is not necessary when using sufficiently advanced optimization methods, as long as overfitting is not the main issue.  Current work shows that the presented method is also able to efficiently train recursive neural networks, with no need to use the structural damping of the Gauss-Newton matrix proposed in~\cite{MartensRNN}.

%

%
%
%
%

\bibliographystyle{plain}
\bibliography{refs}

\end{document}